\documentclass[letterpaper, 10 pt, conference]{ieeeconf}  % Comment this line out if you need a4paper

\IEEEoverridecommandlockouts                              % This command is only needed if 
\usepackage{xspace}
\usepackage{hyperref} 
\usepackage{mathrsfs}
\usepackage{amsfonts}
\usepackage{amsmath}
\usepackage[flushleft]{threeparttable}
\usepackage{tablefootnote}
\usepackage{multirow}
\usepackage{hhline}
\usepackage{graphicx}
\usepackage{subfigure}
\usepackage{hanging}
\usepackage{color}
\usepackage{tikz}
\usepackage{wrapfig}
\usetikzlibrary{calc,arrows,decorations.markings}
\usepackage{balance}
\usepackage{makecell, booktabs}
\usepackage{hhline}
\usepackage{ifthen}
\usepackage{caption}
\usepackage{graphicx}
\usepackage{booktabs}

\let\labelindent\relax
\usepackage{enumitem}

\graphicspath{{../},{../image}}

\newcommand{\netName}{Keypoint-GraspNet\xspace}
\newcommand{\netNameS}{KGN\xspace}

\newcommand{\SE}[1]{SE({#1})}

\newcommand{\trl}{T}
\newcommand{\rotMat}{R}

\newcommand{\graspSet}{\mathcal{G}}
\newcommand{\grasp}{g}
\newcommand{\numGrasp}{N}
\newcommand{\graspWidth}{w}

\newcommand{\ptsSet}{\mathcal{•}{P}}
\newcommand{\pt}{p}
\newcommand{\kptDist}{l}
\newcommand{\gCenter}{c}

\newcommand{\camIntMat}{K}
\newcommand{\RGB}{I}
\newcommand{\Depth}{D}
\newcommand{\height}{H}
\newcommand{\width}{W}

\newcommand{\oriNum}{M}
\newcommand{\gGenFun}{f}

\newcommand{\centerHeatmap}{Y}
\newcommand{\centerOffset}{O}
\newcommand{\centerKptsOffset}{J}
\newcommand{\graspWidthMap}{S}
\newcommand{\threshold}{\epsilon}
\newcommand{\graspNum}{k}

\newcommand{\score}{s}

\newcommand{\loss}{L}

\newcommand{\Real}{\mathbb{R}}
\newcommand{\dist}{d}
\newcommand{\norm}[1]{\|{#1}\|}
\DeclareMathOperator{\Tr}{Tr}

\newcommand{\PnP}[1][]{\ifthenelse{\equal {#1} {}}{P\textit{n}P}{P{#1}P}\xspace}
\newcommand{\EPnP}{EP\textit{n}P}

\title{\LARGE \bf
\netName: Keypoint-based 6-DoF Grasp Generation from the Monocular RGB-D input
}

\author{Yiye Chen$^{1}$, Yunzhi Lin$^{1}$, Ruinian Xu$^{1}$, and Patricio A. Vela$^{1}$%
\thanks{*This work was supported in part by NSF Award \#2026611}%
\thanks{$^{1}$ Y. Chen, Y. Lin, R. Xu, and P.A. Vela are with the
School of Electrical and Computer Engineering, and the
Institute for Robotics and Intelligent Machines, Georgia Institute of
Technology, Atlanta, GA. 
{\tt\small \{yychen2019, ylin466, rnx94, pvela\}@gatech.edu}}%
}

\begin{document}

\maketitle
\thispagestyle{empty}
\pagestyle{empty}

%===============================================================================
% Abstract

\begin{abstract}

%The purpose of this document is to provide both the basic paper template and submission guidelines. Abstracts should be a single paragraph, between 4--6 sentences long, ideally. Gross violations will trigger corrections at the camera-ready phase.
    
% Context 
The success of 6-DoF grasp learning with point cloud input is tempered by
the computational costs resulting from their unordered nature and
pre-processing needs for reducing the point cloud to a manageable size.
These properties lead to failure on small objects with low point cloud
cardinality.
Instead of point clouds, this manuscript explores grasp generation directly
from the RGB-D image input.  The approach, called
\textit{{\netName} {(\netNameS)}}, operates in perception space by detecting
projected gripper keypoints in the image, then recovering their {\SE3} poses
with a {\PnP} algorithm.
% Visual Results
Training of the network involves a synthetic dataset derived from primitive
shape objects with known continuous grasp families.  Trained with only
single-object synthetic data, \textit{\netName} achieves superior result on
our single-object dataset, comparable performance with state-of-art
baselines on a multi-object test set, and outperforms the most competitive
baseline on small objects.  
\textit{\netName} is more than 3x faster than tested point cloud methods.
% Physical Experiments
Robot experiments show high success rate, demonstrating  {\netNameS}'s
practical potential.
Code is available at:~\url{https://github.com/ivalab/KGN}.
\end{abstract}

%===============================================================================
% Introduction

\section{Introduction}\label{sec:Intro}

% The pipeline: Generate, evaluate, and optionally refinement. Emphasis the importance of generating diverse and accurate grasp candidates
Determining grasp configurations from visual sensor input is a
fundamental problem in robot manipulation. While great progress has been
made on the detection of top-down grasps using deep learning
\cite{dexnet22017, deepgrasp2018, gknet2021, CGNet2021}, general purpose object
grasping requires 6-DoF grasp poses for task completion purposes or for
more flexibility in challenging situations. 

A typical 6-DoF grasp system involves 
a \textit{grasp generation} module for generating grasp candidates 
and a \textit{grasp evaluation} module for ranking the candidates
  regarding successful executability based on 
estimated grasp quality \cite{GeoEval2018}, 
collision considerations \cite{collision2021, implicitCollision2021}, 
or 
environemental constraints \cite{humanHandover2021, graspInPrinter2021}.
Some methods include a \textit{grasp pose refinement} module to
enhance the grasp success probability \cite{refineScore2017},
which acts an iterative approach to the grasp synthesis.  
Being the first step, a good \textit{grasp generation} algorithm is a
prerequisite for robust grasping.
An ideal grasp candidate set should contain poses that are accurate and
diverse to ensure functionality and flexibility in constrained environments.

% People are using the point cloud. but the data is unordered, makes extracting the geometric information from it slow, limit the input point number, and usage.
Based on the premise that 6-DoF grasp detection requires 3D geometric
reasoning, the majority of approaces process 3D data.  
Deep point cloud feature extractors, such as PointNet \cite{Pointnet2017}
and PointNet++ \cite{Pointnet++2017}, support grasp detection
from point cloud input \cite{pointnetGPD2019, 6dofGraspNet2019, s4g2020,%
GDN2020, pointnet++grasp2020, ContactGraspNet2021}.
The preference for point clouds is due to their more immediate accessibility
compared to other 3D data formats (e.g., meshes, voxels).
However, geometric information is lost as point clouds are unordered.  
Recovering lost structure involves methods such as hierarchical
grouping and/or geometry-based sampling \cite{Pointnet++2017}. 
Further, point clouds have poor scaling; computational
cost soars as input cardinality increases \cite{Randla-net2020}.
Point cloud grasp detection approaches also involve 
preprocessing to limit point cloud cardinality. 
Methods include segmentation \cite{6dofGraspNet2019, pointnetGPD2019} 
and downsampling \cite{s4g2020}, which may introduce segmentation 
error \cite{ContactGraspNet2021} or discard critical information.
Point cloud methods may produce biased results due to point distribution
imbalance. Observations in \S\ref{sec:visResults} show that small
objects tend to be ignored.

% Here we focus on grasp generation from the RGB-D input.
\paragraph{Contribution}
This paper explores 6-DoF grasp generation directly from monocular RGB-D
(2.5D) input.  
It proposes \textbf{a new single-view grasp generation method, named
\textit{\netName (\netNameS)},} inspired by object pose estimation
\cite{perspectiveNet2019, centerPose2022} and advances in
top-down {\SE2} grasping \cite{gknet2021}.  The method first uses a
convolutional neural network (CNN) to predict the 2D perspective
projection of a set of predefined 3D keypoints in the gripper frame.  A
Perspective-n-Point (\PnP) algorithm then recovers the 3D grasp poses
from established 2D-3D correspondences.
% ADD IN IF ROOM:
% and camera intrinsics.

% Data
\textbf{To generate sufficiently rich and dense training data, we
construct a primitive shape dataset with continuously parameterized and
densely sampled grasp families}. The tactic leverages the simple,
closed-form nature of primitive object geometry to simplify
specification of the feasible grasp pose distribution.  
Once the distribution is defined, sampling involves generating a uniform
cover of the feasible grasp set. This overcomes the sampling
insufficiency of randomized sampling methods and related sample-based
grasp labeling strategies applied to arbitrary objects
\cite{sampleInvest}. 
Using primitive shapes also provides a principled way to evaluate both
accuracy and diversity of the predicted set.  
While primitive shapes may seem limiting, PS-CNN \cite{ps2022} observed
that common household objects are similar to primitive shapes or contain
primitive shape parts.  PS-CNN achieves top 6-DOF grasping performance by
recognizing primitive shape regions. Its drawback is a time-consuming 3D
model fitting process for dense grasp generation.  
Based on these findings, we conjecture that models trained on primitive
shape datasets can extrapolate to realistic objects, which is supported
by physical experiment outcomes in \S \ref{sec:phyExp}.

% Advantage of our method.
Testing shows that {\netNameS}, trained only on single primitive shape
data with a coarse covering of the grasp families, ``fills in'' the
grasp pose gaps and achieves competitive performance.
Specifically, \textbf{{\netNameS} outperforms baselines on a single-object
test set, and achieves comparable results on a multi-object test set to a
baseline specifically trained for cluttered environments.}
It is more than 3-times faster than all baselines, and demonstrates
a significant advantage on small objects compared to point cloud
methods.  Physical experiments verify the transferability of the
approach to the embodied, real-world setting.

%===============================================================================
% Literature

\section{Related Work}\label{sec:Lit}
Here, the focus is on related 3D grasp detection literature.
Top-down grasping literature follows a similar taxonomy.

\paragraph{Grasp Generation}

% First quickly go through the model-based method and the geometric heuristic
Model-based methods estimate the object pose first and then generate
grasps \cite{moped2011}.  Unfortunately, exact object models are not
always available in general purpose grasping.  Alternatively,
GPG \cite{GPG2018} employed a heuristic method based on point cloud
geometry analysis at sampled points. 
Candidate grasp hypotheses at the sampled points depend on the local
surface normal and principle curvature directions. 
Though high quality candidates can be generated and selected with
state-of-art evaluators \cite{pointnetGPD2019, GPD2017}, the time
consumption is large and the diversity of the hypothesis set is limited
(\S\ref{sec:phyExp}). Sampling-based methods suffer from sample sparsity in
the 6-DoF grasp space.

% Then goes for the deep learning based method - First voxel-grid based, then
Deep learning techniques for 3D data and large scale dataset synthesis
\cite{graspnet1b2020, acronym2021} motivate data-driven approaches.
These include methods based on truncated signed distance functions of the
scene using voxels \cite{volumetric2020}, which require multi-view
scanning of the scene; and point cloud sampling methods such as 6-DOF
GraspNet \cite{6dofGraspNet2019} and 
Contact-GraspNet \cite{ContactGraspNet2021}.
%that then regresses the orientation and the open width. Grasp generation
%can propose to encode the feasible grasps as a low dimensional Gaussian
%distribution.
While the grasp generator of \cite{6dofGraspNet2019} was trained as a
Variational Auto-encoder (VAE), \cite{acronym2021} employed a
Generative Adversial Network (GAN) formulation.  Grasp proposals can be
reconstructed from latent space samples.

% TODO: put the end-to-end approaches here
Recent approaches merge the generative and the discriminative components
to result in end-to-end grasping systems. 
The ability of PointNets \cite{Pointnet2017, Pointnet++2017} to extract
local and global feature from point clouds admits candidate-wise quality
regression/classification.
GPNet \cite{GPNet2020} adopts the anchor idea from the object detection research \cite{fasterrcnn2015}. It first generates the initial proposals based on a discrete set of the grid points, then uses a network to verify the antipodal validity \cite{antipodal1993}, regress the approaching direction, and predict the quality.
Several methods \cite{s4g2020, pointnet++grasp2020, L2G2022} predict the per-point grasp confidence and configuration, either in the format of translation and rotation directions \cite{s4g2020, pointnet++grasp2020}, or the two contact points with the pitching angle \cite{L2G2022}.
Extending the idea with discrete orientation classification enables
multiple (coarse to fine) grasp predictions about the same point 
\cite{GDN2020}.

\paragraph{Grasp Evaluation and Refinement}

% the evaluation method
The grasp evaluator aims to eliminate false predictions from the candidate set 
and rank the hypotheses for the selection. 
GPD \cite{GPD2017} projects the local geometry features to the gripper frame planes and trains a CNN to classify the grasp quality. 
PointNetGPD \cite{Pointnet++2017} upgrades the CNN to PointNet \cite{Pointnet2017} and directly processes the point cloud within the gripper closing area.
The GPDs show promising results on heuristically sampled grasps \cite{GPG2018}. 
6DOF-GraspNet \cite{6dofGraspNet2019} follows a similar idea, but represents the candidate as the union of the object and the gripper point sets.
% the refinement
Some work also employs a trained evaluator to improve the grasp poses \cite{6dofGraspNet2019,refineScore2017, refineProb2020}, where the idea is to optimize the estimated quality by refining the initial pose locally within the \SE3 space.
% why we focus on the generation
However, empirical evidence indicates that the performance of a grasp system
depends on the quality of the generated grasps hypotheses, even with a
capable evaluator \cite{6dofGraspNet2019}. Hence, \textbf{this work focuses
on grasp generation instead of grasp evaluation.}

\paragraph{Grasp Detection From the 2D/2.5D input}
Grasp synthesis from point clouds seems reasonable given that solutions
follow the traditional sense-reconstruct-analyze pipeline, while leveraging
point set deep learning models \cite{Pointnet2017, Pointnet++2017}.
However, the increased time costs with point quantity affects runtimes
\cite{Randla-net2020}.  Speed issue stem from the unordered nature of point
cloud data that loses any implicit or explicit geometry.  Offsetting the
loss requires more computation in the form of Farthest Point Sampling (FPS)
and ball-query-based grouping \cite{Pointnet++2017}.  The need to reduce
point cloud cardinality by either segmenting a local point cloud or
downsampling the scene point set incurs additional latency.

RGB-D data, on the other hand, encodes spatial structure that CNN's can be
trained to extract.  Recent CNN-based top-down grasp detection methods
achieve high performance with 2D/2.5D input \cite{gknet2021} based on
network outputs that recover {\SE2} grasps.  The RGBD-Grasp method
\cite{rgbGrasp2021} employs a similar heatmap-based approach to 6-DoF grasp
learning based on discretization of the grasp orientation space, followed by
point-wise processing of top candidates.
\textbf{This paper continues to investigate 6-DoF grasp generation from
2.5D image input.}
It explores keypoints as the intermediate grasp representation to bridge the
dimensionality gap.  Countinuous keypoint coordinates avoid discretization
error and lead to efficient keypoint-based 3D pose recovery.

%===============================================================================
% Problem Definition
%\input{probDef.tex}

%===============================================================================
% Method - 2D-Keypoints-based 6-DoF Grasp Generation

\begin{figure*}[t!]
 
  \vspace*{0.025in}
  \centering
  \scalebox{0.97}{
    \begin{tikzpicture}
     \node[anchor=north west] at (0in,0in)
      {{\includegraphics[width=0.82\textwidth,clip=true,trim=0in
      7.1in 0.1in 0.035in]{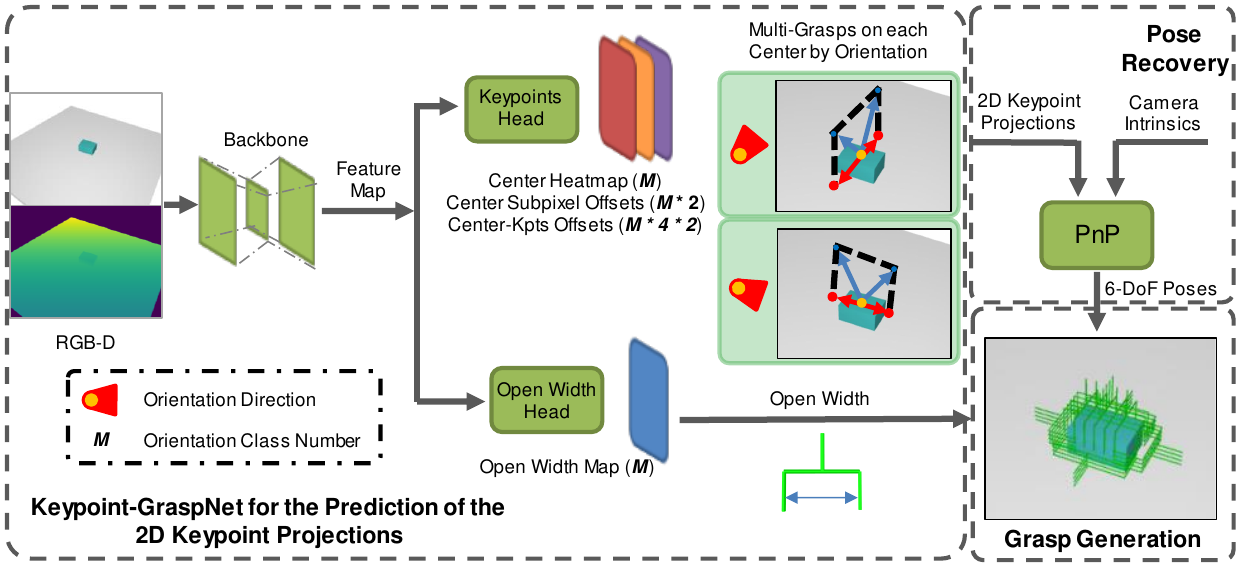}}};
%     \node[yshift=-0pt,anchor=north west] at (0.1in,0.0in) {\bf \small (a)};
%     \node[anchor=north west] at (0.92in,-0.05in) {\textbf{(a)}};
%     \node[anchor=north west] at (2.00in,-0.05in) {\textbf{(b)}};
%     \node[anchor=north west] at (3.09in,-0.05in) {\textbf{(c)}};
    \end{tikzpicture}
  }
  \vspace*{-0.15in}
  \caption{The method pipeline. The bold values in parentheses are channel dimensions.
  \netName\ predicts the probability of each pixel being a grasp center.
  Center prediction is done for each orientation class, to detect multiple
  grasps with a shared center.  For each grasp center, the network regresses
  to subpixel-level the offsets of four image-space keypoints and the grasp
  open widths.  The outputs recover four 2D keypoint projection coordinates,
  called \textit{assistant points}, used to recover 6-DoF grasp poses with a
  \PnP algorithm.
%  The predicted poses and the open widths constitute the grasp generation result $\hat{\graspSet}$.
  }
  \vspace*{-0.2in}
 \label{fig:method}
\end{figure*}

\begin{figure}[t!]
 
  \vspace*{-0.0in}
  \centering
  \scalebox{0.97}{
    \begin{tikzpicture}
     \node[anchor=north west] at (0in,0in)
      {{\includegraphics[width=0.9\linewidth,clip=true,trim=0in
      9.0in 3.9in 0.035in]{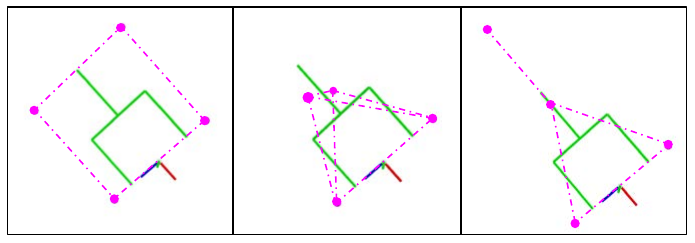}}};
%     \node[yshift=-0pt,anchor=north west] at (0.1in,0.0in) {\bf \small (a)};
     \node[anchor=north west] at (0.85in,-0.05in) {\textbf{(a)}};
     \node[anchor=north west] at (1.85in,-0.05in) {\textbf{(b)}};
     \node[anchor=north west] at (2.87in,-0.05in) {\textbf{(c)}};
    \end{tikzpicture}
  }
  \vspace*{-0.3in}
  \caption{The gripper frame and keypoint options.
  The gripper frame is defined at the finger tips midpoint. 
  \textit{Red}, \textit{green}, \textit{blue} axes represent \textit{x}, \textit{y}, \textit{z} directions, respectively. 
  Keypoints are color-coded as megenta.
  The following keypoint options are considered:
  \textbf{(a)} box-type;
  \textbf{(b)} tetrahedron-type;
  \textbf{(c)} tail-type.
 Option \textbf{(a)} achieves superior performance in (\S\ref{sec:visResults}).
  }
 \label{fig:kpts_opts}
 \vspace{-0.23in}
\end{figure}

\section{Keypoints-Based 6-DoF Grasp Generation}
\label{sec:meth}

%Give an overview of the pipeline: RGB-D to the 2D Keypoint Projections to the 6-DoF grasp poses using the PnP.

\paragraph{Problem Definition\label{sec:ProbDef}}
% Reiterate that we are focusing on the grasp candidate generation.  The output should be a set. 	
Given a single-view color and depth image pair (i.e.  RGB-D), 
$\RGB \in \Real^{\height \times \width \times 3}$ and  
$\Depth \in \Real^{\height \times \width}$, 
the goal is to generate 6-DoF grasp candidates for grasping a visible
target object using a parallel-jaw, or equivalent, gripper.
Equivalently, it is to define the function:
\begin{equation*}
  \graspSet = \{ (\grasp, \graspWidth) \,|\, \grasp \in \SE3, \graspWidth
  \in \Real^+ \} 
            = \gGenFun( \RGB, \Depth ),
\end{equation*}
with grasp pose $\grasp$ and gripper open width $\graspWidth$ outputs.

\paragraph{Keypoint Grasp Representation}
The operation will include an intermediate grasp representation given by
4 keypoint coordinats with pre-defined gripper frame coordinates 
$\pt_g = (\pt_{g}^1, \pt_{g}^2, \pt_{g}^3, \pt_{g}^4)$, where
$\pt_{g}^j \in \Real^3$, $j=1,2,3,4$.
As a first step, a CNN named \netName will generate image space
(projected) keypoint coordinates and open widths:
\begin{equation*}
	\hat{\ptsSet}_{2d} = \{ \hat{\pt}_\RGB^i = (\hat{\pt}^{i1}_\RGB, \hat{\pt}^{i2}_\RGB, \hat{\pt}^{i3}_\RGB, \hat{\pt}^{i4}_\RGB) \}_{i=1}^\numGrasp,
	\quad
	\{\hat{\graspWidth}_i\}_{i=1}^\numGrasp \,,
\end{equation*}
where $\numGrasp$ is the number of the grasp proposals.
The second step applies a \PnP algorithm to recover the gripper frame
with respect to the camera frame, 
\begin{align*}
	\hat{\graspSet} = \{ (\hat{\grasp}_i, \hat{\graspWidth}_i) \,|\, 
      \hat{\grasp}_i = \text{\PnP}( \pt_g, \hat{\pt}_\RGB^i, \camIntMat) 
      \}_{i=1}^N\,,
\end{align*}
given the camera intrinsic matrix $\camIntMat$.
Four keypoints guarantee a solution from the chosen PnP algorithm \cite{ippe}.

\subsection{Grasp Representation}\label{sec:graspRep}

% First introduce the box-type design, and mentioned the other design and refer to the ablation study
Let the 4 keypoints, called \textit{assistant points}, be the corners of a
virtual planar square representing the grasp as depicted in 
Fig.~\ref{fig:kpts_opts}(a). As noted earlier, this point quantity suffices
to provide a unique solution from \PnP algorithms  \cite{ippe, p3p, epnp}. 
The gripper frame is depicted in Fig. \ref{fig:kpts_opts}, such that
the square is in the gripper $x$-$z$ plane with one edge aligning with
the $z$-axis.  Keypoint distances are set to a canonical value $\kptDist$.
Alternative keypoint options include 
a non-planar tetrahedron-type (Fig.\ref{fig:kpts_opts}(b)) and 
a planar tail-type (Fig.\ref{fig:kpts_opts}(c)).
Experiments in \S\ref{sec:visResults} indicate that the box-type exhibits
the best resilience to the noise for pose estimation.

% Then reiterate that the kpts is not on the gripper.
The keypoint distance $\kptDist$ need not equal the grasp open width
$\graspWidth$.  The motivation of this design choice is that when
$\graspWidth$ is small (e.g. the grasps for a thin stick), then \PnP
algorithms applied to a concentrated point set is noise-sensitive.  It
disentangles pose learning from open width learning in the canonical
keypoint distance design.

\subsection{\netName}
\label{sec:network}

%\paragraph{Network Design}
% overview of center-based structure design idea
Fig. \ref{fig:method} depicts the processing flow, its input/output
structure, and the network {\netName} architecture. The trained network
outputs a set of 2D keypoint projection groups $\ptsSet_{2d}$ with 
corresponding open widths $\{\graspWidth\}^\numGrasp$. 
RGB-D inputs have better performance over solely RGB or depth, since
complementary visual and scale information is available.
One challenge is that $\ptsSet_{2d}$ can be dense and involve overlapping
points due to the quantity of grasp candidates per object.
We employ a center point \cite{CenterNet2019, duan2019centernet}, defined to
be the gripper frame origin in Fig.~\ref{fig:kpts_opts}, as the grouping clue.

% Describe the input output
The network detects keypoint groups relative to the center such that the
raw network output includes:
(1) The center heatmap $\hat{\centerHeatmap}$ as the per-pixel possibility of being a grasp center;
(2) The center subpixel offset map $\hat{\centerOffset}$;
(3) The center-to-keypoint offset maps $\hat{\centerKptsOffset}$ for the
displacement between the center and 4 keypoints; and
(4) The open width map $\hat{\graspWidthMap}$.

% How to use the output to get the grasps
At the inference stage, first select top-$\graspNum$ scores from 
$\hat{\centerHeatmap}$ and remove low values with a fixed threshold 
$\threshold_c$, to
obtain center coordinates $\hat{\gCenter}^i = (\hat{\gCenter}^i_x,
\hat{\gCenter}^i_y)$ of the candidate grasp set.  
Obtain keypoint coordinates by adding the center-keypoints offsets to the
center coordinates: $\pt^{ij}_\RGB = \hat{\gCenter}^i +
\hat{\centerKptsOffset}_{\hat{\gCenter}^i j}$, $j=1,2,3,4$, which leads to
the 6-DoF poses using a \PnP algorithm.
Finally, recover the open width prediction by indexing
$\hat{\graspWidthMap}$ at the center location $\hat{\graspWidth}_i =
\hat{\graspWidthMap}_{\hat{\gCenter}^i}$.  The poses and the corresponding
open widths constitute the generated grasp set $\hat{\graspSet}$.

% Address the center-overlaps 
Detecting one grasp per center will omit grasps when multiple occupy the
same.  To address this problem, we detect $\oriNum$ sets of outputs
distinguished by the orientation direction of the gripper \textit{z} axis in
the image space.
Due to the symmetry of the grasp pose, the range of the orientation is
assumed to be $[-\pi/2, \pi/2]$ discretized into $\oriNum$ intervals. 
At the training phase, 
the ground truth outputs are labeled at the corresponding channel index.
The network is trained to produce a set of outputs for each orientation
bin, for up to $\oriNum$ grasp predictions per center pixel.
During inference, candidates with inconsistent orientation angle and class
are discarded.
Though orientation is quantized, keypoint coordinates are continuous in 
$\Real^2$, which avoids quantization error.

\subsection{Training Loss} 
The proposed network is end-to-end trainable with the ground truth outputs,
which can be synthetically generated (see \S\ref{sec:dataset}).
The heatmap branch is trained using the penalty-reduced binary focal loss \cite{focalLoss}:
\begin{align*}
\loss_\centerHeatmap = 
& - \frac{1}{\numGrasp}
\sum_{xym} \\
&
\begin{cases}
	(1 - \hat{\centerHeatmap}_{xym})^\alpha \log (\hat{\centerHeatmap}_{xym}) 
	& 
	\text{if } \centerHeatmap_{xym} = 1 \\
	(1 - \centerHeatmap_{xym})^\beta (\hat{\centerHeatmap}_{xym})^\alpha \log(1 - \hat{\centerHeatmap}_{xym}) 
	& \text{otherwise}
\end{cases}
\end{align*}
with the choices $\alpha=2$ and $\beta=4$ following \cite{CenterNet2019}.
The center subpixel offset loss $\loss_\centerOffset$, the center-keypoint
offset loss $\loss_\centerKptsOffset$, and the open width loss
$\loss_\graspWidthMap$ are formulated as $L_1$ losses on the ground truth
centers.  The total loss is the weight sum of each branch loss:
\begin{align*}
	\loss = \gamma_\centerHeatmap \loss_\centerHeatmap +
    \gamma_\centerOffset \loss_\centerOffset + \gamma_\centerKptsOffset
    \loss_\centerKptsOffset + \gamma_\graspWidthMap \loss_\graspWidthMap.
\end{align*}
The weights used were: $\gamma_\centerHeatmap=1$, $\gamma_\centerOffset=1$, $\gamma_\centerKptsOffset=1$, $\gamma_\graspWidthMap=10$.

%===============================================================================
% Experiment
\section{Experiments}\label{sec:exp}

\begin{table*}[t!]
  \vspace*{0.03in}
  \centering
  \setlength\tabcolsep{2.0 pt} 
  \renewcommand{\arraystretch}{1.3}
  \caption{Vision Dataset Evaluation}
  \begin{threeparttable}
  \begin{tabular}{|c| c | c  c  c  c  c  c  c  c  c | c  c  c  c c c c c c | c|}
  		\hline
	\multirow{2}{*}{Methods} &
	\multirow{2}{*}{Modality} &
    \multicolumn{9}{c|}{\textbf{Single-Object Evaluation (GSR\% / GCR\% / OSR\%) }} &
    \multicolumn{9}{c|}{\textbf{Multi-Object Evaluation (GSR\% / GCR\% / OSR\%) }} &
    \multirow{2}{*}{FPS} \\ 
 		\cline{3-20}
 	{} & {} & \multicolumn{3}{c|}{$1\text{cm}+20^\circ$} &
    \multicolumn{3}{c|}{$2\text{cm}+30^\circ$} & 
    \multicolumn{3}{c|}{$3\text{cm}+45^\circ$} &
    \multicolumn{3}{c|}{$1\text{cm}+20^\circ$} &
    \multicolumn{3}{c|}{$2\text{cm}+30^\circ$} & 
    \multicolumn{3}{c|}{$3\text{cm}+45^\circ$} &
    {} \\\hline
 	
 	PointNetGPD & PC & 
% 	\multirow{2}{*}{\makecell{Point \\ Cloud}} & 
 	% single
 	\multicolumn{3}{c|}{0.43 / 0.13 / 1.50 } &
 	\multicolumn{3}{c|}{1.52 / 0.90 / 3.57 } &
 	\multicolumn{3}{c|}{20.5 / 4.80 / 16.0 } & 
 	% multi
 	\multicolumn{3}{c|}{0.00 / 0.00 / 0.00 } &
 	\multicolumn{3}{c|}{8.33 / 0.02 / 0.80 } &
 	\multicolumn{3}{c|}{41.7 / 0.33 / 3.20 } &  
 	0.008 \\ 
 	
 	\hline
 		
 	6DoF-GraspNet & PC & 
 	 	% single
 	\multicolumn{3}{c|}{3.78 / 6.78 / 35.4} &
 	\multicolumn{3}{c|}{16.5 / 39.6 / 79.1} &
 	\multicolumn{3}{c|}{35.9 / \textbf{73.9} / 97.7} & 
 	% multi
 	\multicolumn{3}{c|}{0.20 / 0.10 / 0.70} &
 	\multicolumn{3}{c|}{2.00 / 0.50 / 5.27} &
 	\multicolumn{3}{c|}{8.66 / 2.68 / 16.7} &  
 	0.240 \\ 
 	
	\hline
	\hline

 	Contact-Graspnet\tnote{1} & PC &
 	 	% single
 	\multicolumn{3}{c|}{29.9 / 24.9 / 77.0 } &
 	\multicolumn{3}{c|}{60.1 / 32.0 / 81.7 } &
 	\multicolumn{3}{c|}{81.6 / 36.5 / 84.2} & 
 	% multi
 	\multicolumn{3}{c|}{\textbf{22.1} / \textbf{15.5} / \textbf{44.1}} &
 	\multicolumn{3}{c|}{\textbf{54.2} / \textbf{28.5} / 51.4} &
 	\multicolumn{3}{c|}{\textbf{78.4} / \textbf{34.5} / 54.4} &  
 	\thead{2.109}
 	\\
 	
 	\hline
	\hline
	
	RGB-Matters	& RGB-D & 
 	\multicolumn{3}{c|}{ 0.74 / 0.12 / 1.9  } &
 	\multicolumn{3}{c|}{ 4.68 / 1.30 / 8.5 } &
 	\multicolumn{3}{c|}{ 21.92 / 6.13 / 20.6 } & 
 	% multi
 	\multicolumn{3}{c|}{ 1.36 / 0.22 / 1.65 } &
 	\multicolumn{3}{c|}{ 6.33 / 1.67 / 5.88 } &
 	\multicolumn{3}{c|}{ 26.40 / 6.46 / 16.73 } & 
 	0.409
 	\\

 	\hline
 		
 	\textbf{\netNameS}	& RGB-D & 
 	\multicolumn{3}{c|}{\textbf{55.5} / \textbf{42.9} / \textbf{97.0} } &
 	\multicolumn{3}{c|}{\textbf{78.5} / \textbf{63.3} / \textbf{99.6} } &
 	\multicolumn{3}{c|}{\textbf{86.9} / 73.2 / \textbf{99.9}} & 
 	% multi
 	\multicolumn{3}{c|}{10.8 / 5.48 / 28.7} &
 	\multicolumn{3}{c|}{30.6 / 18.7 / \textbf{51.8}} &
 	\multicolumn{3}{c|}{49.6 / 33.8 / \textbf{62.4}} & 
 	\textbf{9.290}
 	\\
 		\hline
  \end{tabular}
  \begin{tablenotes}
	  \item[1] Contact-GraspNet is trained on cluttered scenes as opposed to other methods. 
	  In addition, ground truth object segmentation mask is provided to cluster input point clouds.
	  Hence, it serves as a upper bound for the multi-object evaluation.
  \end{tablenotes}
  \end{threeparttable}
  \label{tab:visResults}
  \vspace*{-0.1in}
\end{table*}

\subsection{Dataset Generation}
\label{sec:dataset}

% What is primitive shape and grasp family
The ground truth training annotations are generated from known grasp poses
and camera poses.  To train and evaluate \netNameS, we generate a synthetic
dataset based on the primitive shape and grasp family idea \cite{ps2022}. 
A grasp family is a pre-defined parametric grasp pose space for a shape
primitive instance.  With the continuous, closed-form grasp pose
descriptions, uniformly sample the grasps and create a set covering based on
spatial and angular distance thresholds.

% detail
Six shape categories are used \cite{ps2022}: \textit{Cylinder}, \textit{Ring}, \textit{Stick}, \textit{Sphere}, \textit{Semi-sphere}, \textit{Cuboid}.
For single-object scenes, randomly choose a shape class, and generate an
object of random size and color.
Place the object on a table with a random stable pose.
For grasp annotations, sample object-frame grasp poses from uniform
set coverings of the pre-defined grasp families.
Grasps causing collision between a virtual gripper model and the table or
other objects are removed. 
For multi-object data, the procedure is repeated to place one object per
shape in the scene.
Finally, 5 virtual camera poses are sampled for RGB-D image rendering.
We generate $1000$ single-object scenes following the above procedure, and
divide them into an $80\%$/$20\%$ training/testing split. 
We also create, $200$ multi-object test scenes to examine \netNameS's
generalization ability to cluttered test cases.
In total, $4000$ single-object training , $1000$ single-object testing , and
$1000$ multi-object testing data instances are generated.

\textbf{Rationale behind primitive shapes: }
We first argue for the benefit of primitive shapes for evaluation.
Evaluating both the accuracy and comprehensiveness of a predicted set requires annotations with sufficient coverage of grasp space \cite{deepgrasp2018, 6dofGraspNet2019, ContactGraspNet2021}.
Recent literature \cite{jacquard2018,graspnet1b2020, acronym2021} adopts a sample-then-verify process done in simulation.
However, sampling leads to biased or incorrect results \cite{sampleInvest, gknet2021}, 
which negatively impacts training and evaluation.

Sampling problems are caused by the fact that the grasp distribution for
an arbitrary (non-primitive) object is complex and has no explicit formula.
To tackle the challenge, we restrict the objects to primitive shapes. 
Due to the simplicity of the object geometry, closed-form expressions of
the object's grasp families sufficiently cover the feasible grasp modes.  
The closed-form nature of the grasp families permits deterministic
annotation sampling with a desired density.

Lastly, it is observed that primitive shapes comprise a wide range of common
household objects.  State-of-art performance has been achieved by explicitly
targeting basic geometrical regions \cite{ps2022}.

\textbf{Extrapolation from sparse to dense:}
Another challenge is the gap between the continuous grasp space and the
discrete labels for training.
To investigate the effect of training with sparse annotations, we
vary the sample density between the training and testing split.
The training set uniformly samples $5$ and $11$ elements from the 
\textit{translational freedom} and \textit{rotational freedom}, respectively, 
whereas the testing set samples $10$ and $30$ elements.
The increased density (and stricter error threshold) at the evaluation stage
examines the network's ability to learn the underlying distribution.

\subsection{Implementation Details}
\label{sec:detail}

The backbone network is DLA-34 \cite{dla} with deformable convolution
layers \cite{dcn}, designed by \cite{CenterNet2019}, which produces feature
maps one-fourth the input dimensions.
Each task head is a shallow network of a $3\times3$ convolution layer with
$256$ channels followed by a $1\times1$ convolution layer to the desired
channel dimension.  There are $\oriNum=9$ orientation classes.

The network training data is the training split dataset described in 
\S \ref{sec:dataset}.  The labeled grasps are conditionally flipped (rotated
by $180^{\circ}$ around the $x$ axis) to consistently point the gripper $y$
axis to the left in the image space.  Only one grasp per orientation class
per center pixel is kept.  The input image data is augmented with random
cropping, flipping, and color jittering.

Network training uses the Adam optimizer for 400 epochs. The initial
learning rate is $1.25e^{-4}$, reducing by 10x at epochs 350 and 370. For all experiments, the center heatmap threshold at the inference phase is fixed to $\threshold_c = 0.3$, and the top-$k$ number is set to $k = 100$.

\begin{figure*}[t!]
  \vspace*{0.0in}
  \centering
  \scalebox{0.98}{
    \begin{tikzpicture}
     \node[anchor=north west](figs) at (0in,0in)
      {{\includegraphics[width=0.85\linewidth,clip=true,trim=0in
      7.9in 0.1in 0.035in]{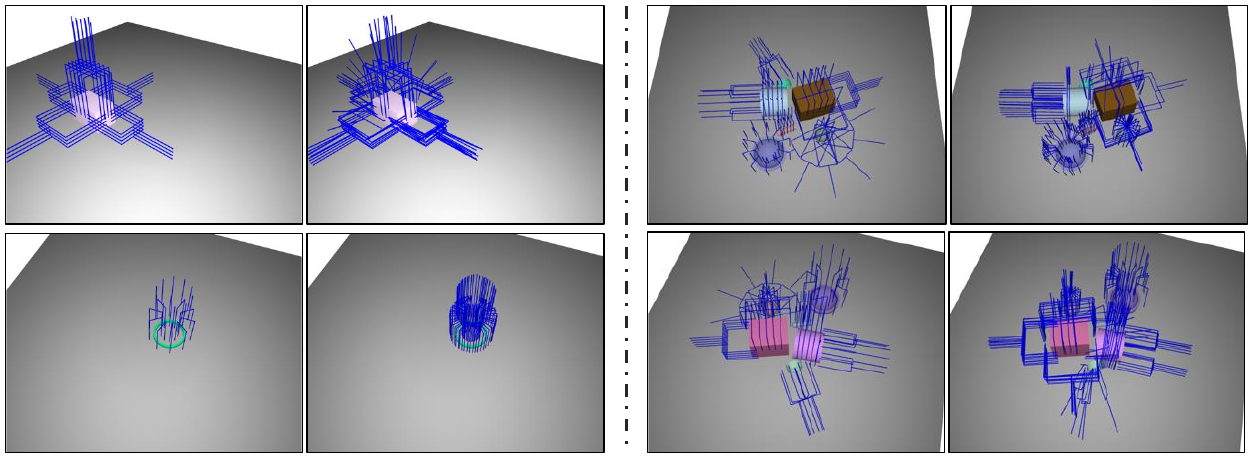}}};
%     \node[yshift=-0pt,anchor=north west] at (0.1in,0.0in) {\bf \small (a)};
     \node[anchor=north] at ($ (figs.south) + (-2.25in, 0.07in) $) {\bf
     \small Training Labels};
     \node[anchor=north] at ($ (figs.south) + (-0.8in, 0.07in) $) {\bf
     \small \netNameS};
     \node[anchor=north] at ($ (figs.south) + (0.8in, 0.07in) $) {\bf \small
     Training Labels};
     \node[anchor=north] at ($ (figs.south) + (2.25in, 0.07in) $) {\bf
     \small \netNameS};
%     \node[anchor=north west] at (0.9in,-1.9in) {Keypoints \\ Prediction};
    \end{tikzpicture}
  }
  \vspace*{-0.1in}
  \caption{Example results on synthetic single-object and multi-object datasets.
  Trained only on single-object data with sparse annotations, 
  \netName\ predicts dense grasp poses (left) and extrapolates to
  multi-object scenarios (right).
  }
 \label{fig:visResults}
 \vspace*{-0.2in}
\end{figure*}

\subsection{Synthetic Dataset Experiment Setup}

\label{sec:visExpSetup}

Evaluation with the test splits of the synthetic dataset uses several metrics.
Three point cloud methods are compared with as baselines.  
%Furthermore, we ablate on the keypoint and {\PnP} algorithm choices.

%\subsubsection{Metrics and Baselines}
%\label{sec:metric}

\textbf{Metrics: }
The metrics quantify when two grasp poses are similar.
Two $\SE3$ elements $\grasp$ and $\grasp^*$ are similar up to the threshold 
$\threshold_\trl$, $\threshold_\rotMat$ if their translational and
rotational components satisfy:
\begin{align*}
	&\dist_\trl (\trl, \trl^*) = \norm{\trl - \trl^*}_2 \leq \threshold_\trl \\
	&\dist_\rotMat (\rotMat, \rotMat^*) = \arccos( \frac{1}{2} \Tr(\rotMat \rotMat^*) - \frac{1}{2})) \leq \threshold_\rotMat
\end{align*}
where $\trl$ and $\rotMat$ represent the translation vector and rotation
matrix from an $\SE3$ element. 
The rotation distance is the minimum angle required to align two rotations \cite{localBenchmark, hartley2013rotation}.

Given a set of predicted grasps $\graspSet$ and ground truth grasps $\graspSet^*$, three metrics evaluate the predictions: 
\textit{grasp success rate (GSR)} \cite{6dofGraspNet2019}, 
\textit{grasp coverage rate (GCR)} \cite{6dofGraspNet2019}, and 
\textit{object success rate (OSR)} \cite{deepgrasp2018, gknet2021}. 
The \textit{GSR} is the percentage of the successful predictions to ground truth
grasps (i.e., the predicted
grasp is similar to a ground truth grasp); 
The \textit{GCR} is the percentage of the ground truth grasps similar to at least
one of predicted grasp; 
The \textit{OSR} is the percentage of objects with one or more successful
predictions. 
The three metrics measure the accuracy, the diversity, and the
practicality of the predicted set.

The visual experiments record the metrics under three distance threshold
levels from strict to loose:
$(\epsilon_{\trl}, \epsilon_{\rotMat}) = (1 \text{cm}, 20^{\circ})$, $(2 \text{cm}, 30^{\circ})$,
and $(3 \text{cm}, 45^{\circ})$.

\textbf{Baselines: }
Comparison is with three point cloud grasp generation baselines, namely
\textit{PointNetGPD} \cite{pointnetGPD2019}, \textit{6DoF-GraspNet}
\cite{6dofGraspNet2019}, and \textit{Contact-GraspNet}
\cite{ContactGraspNet2021}; 
and one RGB-D method \textit{RGB-Matters} \cite{rgbGrasp2021}, 
The first three choices cover distinct pipelines:
heuristic grasp sampling followed by ranking with learnt evaluator \cite{pointnetGPD2019};
generator-based grasp synthesis selected and refined by a trained evaluator \cite{6dofGraspNet2019};
and the per-point grasp scoring and parameter regression \cite{ContactGraspNet2021}.
For these baselines, we remove the tabletop points to reduce the input point
cloud cardinality, which is not required by {\netNameS}.
The generator of 6DoF-GraspNet is trained from scratch on our training set.
Pretrained weights are used for the other modules from the baselines due to
their hardware requirements.  The workstation used has an NVIDIA GTX 1080
GPU and an Intel i7-7700 CPU @ 3.60GHz.

\subsection{Synthetic Dataset Experiment Results}
\label{sec:visResults}

\textbf{Single-Object Evaluation: }
Results from {\netNameS} and the baselines on the single-object test set are
shown in Table \ref{tab:visResults}.
{\netNameS} outperforms the baselines under all distance thresholds and metrics except for the GCR at the largest error tolerance.
The grasp success rate gap is consistantly over $45\%$, and the grasp coverage rate gap is over $30\%$ at the strictest error tolerance level.
Although trained with a coarse set covering for each grasp family, the high
GCR outcomes with a denser set covering indicates that {\netNameS} captures the underlying grasp distribution shown in Fig. \ref{fig:visResults}.

\textbf{Multi-Object Evaluation: }
Next {\netNameS} \textit{trained only with single-object data} is applied to
the multi-object test set.
We observe that the predicted \textit{scale} (i.e. translation magnitude
along camera projection) deteriorates in this setting,
as the keypoints proximity in the image space is unstable in the presence of
visual disturbance.
To compensate, we added a depth-based scale refinement:
$\trl = \trl \norm{\trl}_2 / \Depth_{\gCenter}$,
where $\trl$ is predicted camera-to-grasp translation, and 
$\Depth_{\gCenter}$ is the depth image value at predicted grasp center.
Scale prediction is an known deficiency in monovular pose estimation from 2D
image points. Exploration is needed on how the depth channel can best
help to resolve scale.
Meanwhile, Table \ref{tab:visResults} collects the test results. 
{\netNameS} continues to outperform PointNetGPD and 6DoF-GraspNet.
It is second best to Contact-GraspNet, which is \textit{explicitly
trained for cluttered environments}. 
Factors influencing the {\netNameS}
outcomes are the fixed confident threshold and upper limit on number of grasp predictions 
for more objects.

\textbf{Advantage for Small Objects: }
Although Contact-GraspNet outperforms {\netNameS} in the multi-object test,
{\netNameS} is more effective on small objects.
Table \ref{tab:smallObjs} reports the GCR and OSR for the two smallest
objects, namely the \textit{stick} and the \textit{ring}, which have the
the smallest number of occupied observable pixels on average. 
{\netNameS} performs better than Contact-GraspNet.  A low OSR indicates that
the point cloud method tends to ignore small objects due to relative paucity of
points versus larger objects. 

\subsection{Abalation Study on Keypoint Options}
\begin{figure}[t!]
  %\vspace*{-0.15in}
  \centering
  \scalebox{0.98}{
    \begin{tikzpicture}
     \node[anchor=north west] at (0in,0in)
      {{\includegraphics[width=0.88\linewidth,clip=true,trim=0in 0in 0in
      0.5in]{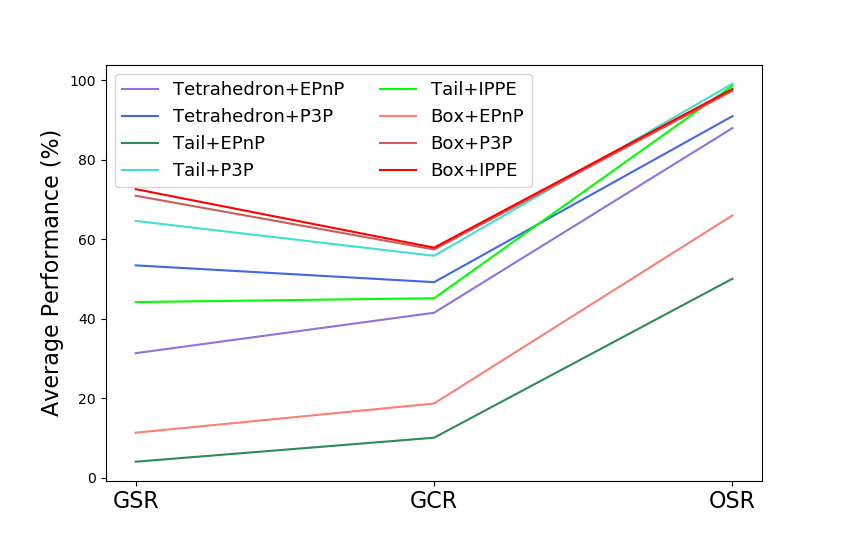}}};
%     \node[yshift=-0pt,anchor=north west, fill=white, inner sep=1.4pt] at (0.07in,-0.05in) {\bf \small (a)};
%     \node[yshift=-0pt,anchor=north west, fill=white, inner sep=1.4pt] at (1.44in,-0.05in) {\bf \small (b)};
%     \node[yshift=-0pt,anchor=north west, fill=white, inner sep=1.4pt] at (2.80in,-0.05in) {\bf \small (c)};
%     \node[yshift=-0pt,anchor=north west, fill=white, inner sep=1.4pt] at (4.16in,-0.05in) {\bf \small (d)};
    \end{tikzpicture}
  }
  \vspace*{-0.2in}
  \caption{
  Ablation study on keypoint and \PnP options with single-object evaluation.
  Numbers are averaged over all error levels.
  The Box+IPPE combination yields the best results.
  }
 \label{fig:ablKpts}
 \vspace*{-0.20in}
\end{figure}

%\textbf{Ablation on the keypoint options: }
Testing across keypoint grouping geometry and {\PnP} algorithm options on
the single-object test set in Sec.\ref{sec:visResults} leads to the outcomes
in Fig. \ref{fig:ablKpts}.  
Per Fig. \ref{fig:kpts_opts}, the keypoint types are: box, tetrahedron, and 
tail. The {\PnP} algorithms tested are:
\PnP[3] \cite{p3p}, \EPnP \cite{epnp}, and IPPE \cite{ippe}.
The results indicate that the combination of planar box-type keypoints
with the IPPE algorithm is the best option. 

%(1) \PnP[3] \cite{p3p}, 
%(2) \EPnP \cite{epnp}, 
%and (3) IPPE \cite{ippe}.
%\PnP[3] is a algebraic approach, 
%\EPnP\ infers the 3D coordinates in the camera frame utilizing the control points, 
%and IPPE is designed to overcome noise for the pose estimation with only coplanar points.

%(1) algebraic approach \PnP[3] \cite{p3p}, 
%(2) control-points approach \EPnP \cite{epnp}, 
%and (3) IPPE \cite{ippe} for the noise resistance for pose estimation with coplanar points.

\subsection{Advantage Highlights}
We summarize the benefits of our RGB-D approach over point cloud methods
observed from the experiments:

\textbf{Preprocessing free: } 
\netNameS directly processes the input data.  In contrast, point cloud
methods require preprocessing such as background removal to reduce input
point number.

\textbf{Speed: } 
Although all baselines have an advantage from excluding segmentation
timing, {\netNameS} still achieves the lowest times and is over 3x faster
than point cloud methods using pre-processed inputs.

\textbf{Superior performance on small objects: } 
Point cloud approaches tend to have poor performance on small objects due to
point cloud paucity.  With the aid of RGB visual cues, \netNameS achieves
higher performance.

\begin{table}[t!]
  \vspace*{0.03in}
  \centering
  \caption{Performance on Small Objects}
  \renewcommand{\arraystretch}{1.2}

  \begin{threeparttable}
  \begin{tabular}{|c| c  c | c  c |}
      \hline
	\multirow{2}{*}{Methods} & 
	\multicolumn{2}{c|}{Stick} & 
	\multicolumn{2}{c|}{Ring} \\
  		\cline{2-5}
	{} & 
	\multicolumn{2}{c|}{${}^\star$Avg GCR\% / OSR\%} &
	\multicolumn{2}{c|}{${}^\star$Avg GCR\% / OSR\%} \\
 		\hline
 	Contact-GraspNet & 
 	\multicolumn{2}{c|}{13.8 / 7.10} & 
 	\multicolumn{2}{c|}{16.1 / 15.8} \\
 	
 	\netNameS & 
 	\multicolumn{2}{c|}{\textbf{37.9} / \textbf{22.2}} & 
 	\multicolumn{2}{c|}{\textbf{45.7} / \textbf{36.3}} \\
 	
    \hline
  \end{tabular}
  \begin{tablenotes}
	  \item ${}^\star$ Numbers averaged over three error tolerance levels.
  \end{tablenotes}
  \end{threeparttable}
  \label{tab:smallObjs}
  \vspace*{-0.15in}
\end{table}

%\begin{table}[t!]
%  \vspace*{-0.0in}
%  \centering
%  
%%  \setlength\tabcolsep{5pt} 
%  \renewcommand{\arraystretch}{1.1}
%   \caption{Physical Experiment Results}
%  \begin{threeparttable}
%  \begin{tabular}{|c| c | c | c |}
%  		\hline
%	Object & Succ & Object & Succ \\
% 		\hline
% 	Ball & 4/5 & Cable & 5/5 \\
% 	Tape & 5/5 & Box & 5/5 \\
% 	Bowl & 4/5 & Mug & 4/5 \\
% 	ToothBrush & 5/5 & Clamp & 3/5 \\
% 		\hhline{|=|=|=|=|}
% 	Mean & \multicolumn{3}{c|}{87.5\%} \\
%		\hline
%  \end{tabular}
%  \label{tab:phyResults}
%  \end{threeparttable}
%%  \vspace*{-0.6in}
%\end{table}

%\begin{figure}
%\begin{floatrow}
%\ffigbox{%
%  \scalebox{0.97}{
%     \begin{tikzpicture}
%      \node[anchor=north west] at (0in,0in)
%       {{\includegraphics[width=0.5\textwidth,clip=true,trim=0in
%       4.3in 0.in 0.in]{rebuttal/figPhyObjSet.pdf}}};
%     \end{tikzpicture}
%   }
%   \vspace*{-0.2in}
%}{%
%  \caption{A figure}%
%}
%\capbtabbox{%
%   \renewcommand{\arraystretch}{1.1}
%   \caption{Physical Experiment Results}
%  \begin{threeparttable}
%  \begin{tabular}{|c| c | c | c |}
%  		\hline
%	Object & Succ & Object & Succ \\
% 		\hline
% 	Ball & 4/5 & Cable & 5/5 \\
% 	Tape & 5/5 & Box & 5/5 \\
% 	Bowl & 4/5 & Mug & 4/5 \\
% 	ToothBrush & 5/5 & Clamp & 3/5 \\
% 		\hhline{|=|=|=|=|}
% 	Mean & \multicolumn{3}{c|}{87.5\%} \\
%		\hline
%  \end{tabular}
%  \label{tab:phyResults}
%  \end{threeparttable}
%}{%
%  \caption{A table}%
%}
%\end{floatrow}
%\end{figure}

\begin{table}
    \vspace{0.03in}
	\begin{minipage}[]{0.54\linewidth}
        \vspace*{-0.02in}
		\setlength\tabcolsep{2.5pt} 
  		\renewcommand{\arraystretch}{1.1}
  		\begin{threeparttable}
  			\caption{Physical Experiment Results}
  			\begin{tabular}{|c| c | c | c |}
  				\hline
				Object & Succ & Object & Succ \\
 				\hline
 				Ball & 4/5 & Cable & 5/5 \\
 				Tape & 5/5 & Box & 5/5 \\
 				Bowl & 4/5 & Mug & 4/5 \\
 				ToothBrush & 5/5 & Clamp & 3/5 \\
 				\hhline{|=|=|=|=|}
 				Mean & \multicolumn{3}{c|}{87.5\%} \\
				\hline
  			\end{tabular}
  			\label{tab:phyResults}
  		\end{threeparttable}
	\end{minipage}\hfill
	\begin{minipage}[]{0.45\linewidth}
		\centering
		\vspace*{0.05in}
		{\includegraphics[width=0.95\textwidth]{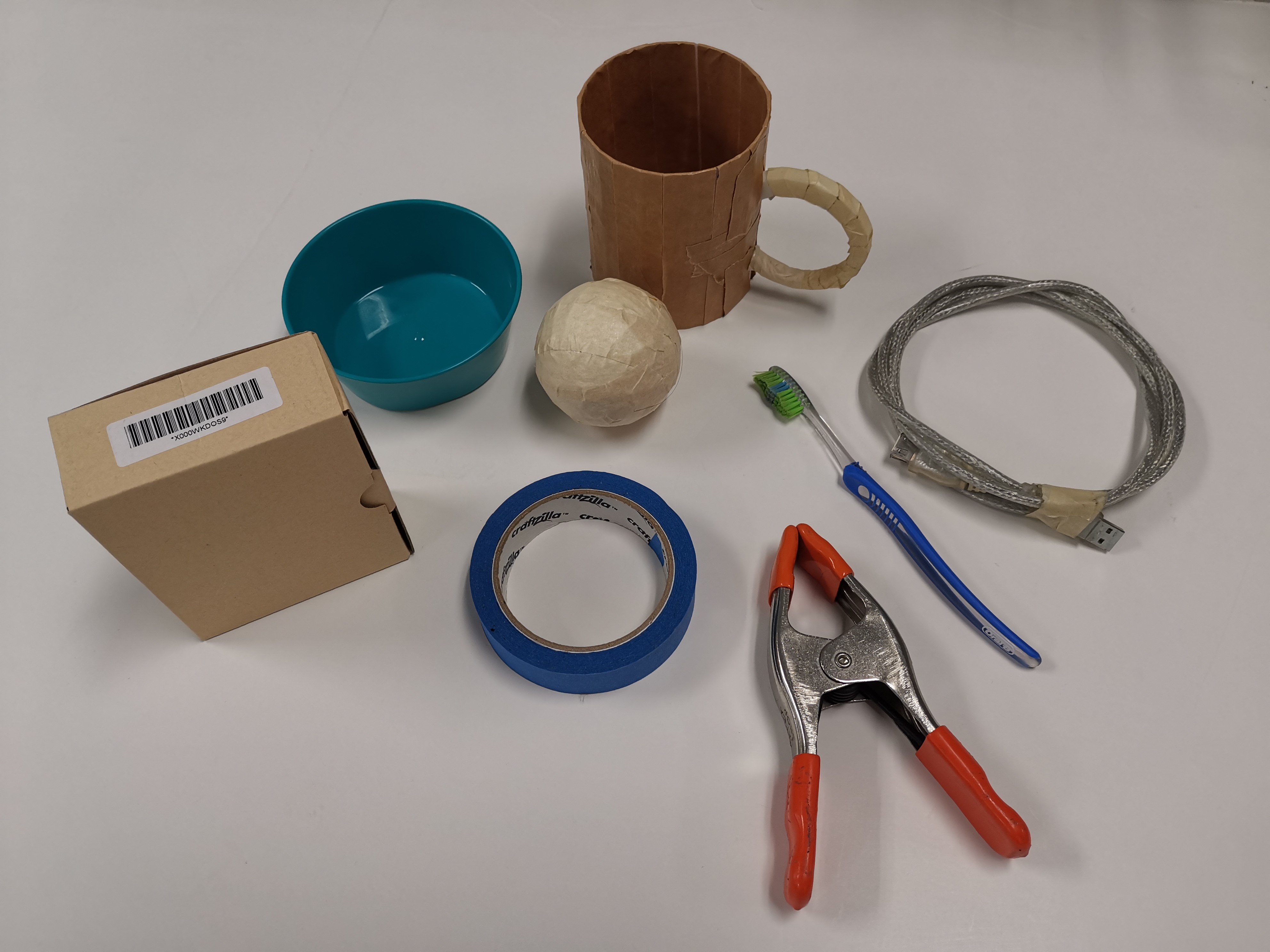}}
		\captionof{figure}{Physical object set}
		\label{fig:phyObjs}
	\end{minipage}
	\vspace{-0.15in}
\end{table}

\subsection{Physical Experiments}
\label{sec:phyExp}

Physical experiments are conducted to validate the grasp generation approach
under the covariate shift and real-world sensor measurements. They involve
single object pick-and-place experiments, where the goal is to pick up a
randomly placed target and drop it at the designated location.
A trial is successful if the object is moved to the spot.
The \textit{success rate}--defined as the ratio of the successful trials to
total trials--is the evaluation criteria.  We adopt 8 common household
objects that covering the primitive shape classes as shown in Fig.
\ref{fig:phyObjs}.  For each object, 5 trials are performed.

A simple grasp candidate ranking and refinement process is employed. 
For the ranking, we score the candidates combining the center confidence in the keypoint detection stage and the reprojection error (RE) induced by the pose recovery stage:
$\score(\grasp) = \centerHeatmap_{xym} + RE$.
The RE, defined as the image distance between the input and estimated 2D keypoints, quantifies the uncertainty of the \PnP algorithm.
We refine the translation scale based on the observed depth from the depth
image as per Sec. \ref{sec:visResults}.

The physical experiment results are shown in Table \ref{tab:phyResults}. 
Trained on a simple synthetic dataset, {\netNameS} achieves satisfying
success rate on real objects, indicating its potential. For comparison,
6-DOF GraspNet had an 88\% success rate for simple objects (box, cylinder,
bowl, and mug) similar to {\netNameS}, but involved more extensive and
computationally involved physics-based simulation.  The primitive shapes
with continuously parametrized models achieve similar performance with a
fraction of the effort and with a faster runtime.

%Example 6-DoF grasp candidates on the objects of different poses are illustrated in Fig. \ref{fig:phyResults}. 

%It is able to produce different grasping strategies is the $\SE3$ depending on the object pose. \ycREF{An figure of different grasp poses on the lying and standing tape.}
%Failure cases involves attempting to grasp short object in a parallel-to-table way, causing the collision between the gripper and the table, 
%and recognizing the colored subpart of an object as a graspable target due to the uniform object color setting in our training dataset.

%\input{tables/tabPhyResults.tex}

%===============================================================================
% Conclusion
\section{Conclusion}
\label{sec:conclusion}

This paper investigated a keypoint-based solution to the problem of 6-DoF grasp
generation from RGB-D input.  Named \textit{\netName (\netNameS)} the
proposed solution detects image space gripper keypoint projections from
which it recovers the 6-DoF grasp pose using a \PnP algorithm.
On a synthesized primitive shape dataset, {\netNameS} is shown to generate
diverse and accurate grasp candidates. It improves grasp performance
for small objects and has lower runtime costs compared to the selected
baselines.  Physical experiments show that the trained model applies to
real-world sensors and manipulators without further finetuning.

Improvements can be made to this proof-of-concept work.  The first is to
explore pose scale estimation more robust to novel visual contents and
object spacing.  Furthermore, the PS dataset labels grasp poses in a
gripper-agnostic manner to examine \netNameS's abilty to approximate
distribution in \SE3 space.  Future work should explore modifying the
training dataset with labels verified by simulation for specific gripper
geometry.  Lastly, applying KGN from multiple views might improve the
grasp pose accuracy around occluded areas.

\bibliographystyle{IEEEtran}
%\balance
\bibliography{reference.bib}

\end{document}